\documentclass[lettersize,journal]{IEEEtran}
\usepackage{amsmath,amsfonts}
\usepackage{algorithm}
\usepackage{array}
\usepackage[caption=false,font=normalsize,labelfont=sf,textfont=sf]{subfig}
\usepackage{textcomp}
\usepackage{stfloats}
\usepackage{url}
\usepackage{verbatim}
\usepackage{graphicx}
\usepackage{cite}
\usepackage{balance}

\usepackage{algorithm}
\usepackage{algpseudocode}
\usepackage{amsmath,amssymb}

\usepackage{booktabs}
\usepackage{multirow}

\newcommand{\para}[1]{\vspace{2pt}\noindent\textbf{#1.~}}
\newcommand{\ignore}[1]{}
\newcommand{\system}{\sloppy{VLAMotor\@}}

\usepackage{ulem}
\usepackage[switch]{lineno}

\usepackage{listings, xcolor}
\definecolor{verylightgray}{rgb}{.97,.97,.97}
\lstdefinelanguage{Solidity}{
  keywords=[1]{anonymous, assembly, assert, balance, break, call, callcode, case, catch, class, constant, continue, constructor, contract, debugger, default, delegatecall, delete, do, else, emit, event, experimental, export, external, false, finally, for, function, gas, if, implements, import, in, indexed, instanceof, interface, internal, is, length, library, log0, log1, log2, log3, log4, memory, modifier, new, payable, pragma, private, protected, public, pure, push, require, return, returns, revert, selfdestruct, send, solidity, storage, struct, suicide, super, switch, then, this, throw, transfer, true, try, typeof, using, value, view, while, with, addmod, ecrecover, keccak256, mulmod, ripemd160, sha256, sha3}, 
  keywordstyle=[1]\color{blue}\bfseries,
  keywords=[2]{address, bool, byte, bytes, bytes1, bytes2, bytes3, bytes4, bytes5, bytes6, bytes7, bytes8, bytes9, bytes10, bytes11, bytes12, bytes13, bytes14, bytes15, bytes16, bytes17, bytes18, bytes19, bytes20, bytes21, bytes22, bytes23, bytes24, bytes25, bytes26, bytes27, bytes28, bytes29, bytes30, bytes31, bytes32, enum, int, int8, int16, int24, int32, int40, int48, int56, int64, int72, int80, int88, int96, int104, int112, int120, int128, int136, int144, int152, int160, int168, int176, int184, int192, int200, int208, int216, int224, int232, int240, int248, int256, mapping, string, uint, uint8, uint16, uint24, uint32, uint40, uint48, uint56, uint64, uint72, uint80, uint88, uint96, uint104, uint112, uint120, uint128, uint136, uint144, uint152, uint160, uint168, uint176, uint184, uint192, uint200, uint208, uint216, uint224, uint232, uint240, uint248, uint256, var, void, ether, finney, szabo, wei, days, hours, minutes, seconds, weeks, years},  
  keywordstyle=[2]\color{teal}\bfseries,
  keywords=[3]{block, blockhash, coinbase, difficulty, gaslimit, number, timestamp, msg, data, gas, sender, sig, value, now, tx, gasprice, origin},  
  keywordstyle=[3]\color{violet}\bfseries,
  identifierstyle=\color{black},
  sensitive=false,
  comment=[l]{//},
  morecomment=[s]{/*}{*/},
  commentstyle=\color{gray}\ttfamily,
  stringstyle=\color{red}\ttfamily,
  morestring=[b]',
  morestring=[b]"
}
\usepackage{pifont}
\usepackage{oplotsymbl}
\usepackage{siunitx}
\usepackage{array,framed}
 \usepackage{
   color,
   float,
   epsfig,
   wrapfig,
   graphics,
   graphicx,
   subcaption
 }
\usepackage{setspace}
\usepackage{amsfonts}
\usepackage{latexsym,fancyhdr,url}
\usepackage{enumerate}
\usepackage{algpseudocode}
\usepackage{graphics}
\usepackage{xparse} 
\usepackage{xspace}
\usepackage{multirow}
\usepackage{microtype}

\usepackage{fancyhdr}

\begin{document}

\title{VLAMotor: Test-Guided Enhancement of Vision-Language-Action Models via Agent-Based Data Synthesis}

\author{
Zeqin Liao, 
Peifan Ren, 
Zixu Gao, 
Hongyu Gong,
Lianyu Hu*,
Wenbing Tang,
Yuhong Nan,~\IEEEmembership{Member,~IEEE},
Zibin Zheng,~\IEEEmembership{Fellow,~IEEE},  
Yang Liu,~\IEEEmembership{Senior Member,~IEEE}
\thanks{ 
\setlength{\parindent}{0pt}
Zeqin~Liao, Lianyu~Hu and Yang~Liu are with School of computing and data science, Nanyang Technological University.
Peifan~Ren, Zixu~Gao, Hongyu~Gong, Yuhong~Nan, and Zibin~Zheng are with School of Software Engineering, Sun Yat-sen University, China and GuangDong Engineering Technology Research Center of Blockchain, China. 
Wenbing~Tang is with Northwest A\&F University,

E-mail: \{zeqin.liao, LianyuHu, YangLiu \}@ntu.edu.sg

E-mail: \{renpf, gaozx9, gonghy8 \} @mail2.sysu.edu.cn

E-mail: \{nanyh, zhzibin\}@mail.sysu.edu.cn

E-mail: \{wenbingtang@nwafu.edu.cn \} 

Lianyu Hu is the corresponding author.
}



}

\markboth{Journal of \LaTeX\ Class Files,~Vol.~14, No.~8, August~2021}%
{Shell \MakeLowercase{\textit{et al.}}: A Sample Article Using IEEEtran.cls for IEEE Journals}


\maketitle

\begin{abstract}
  
Vision-Language-Action (VLA) models follow a data-driven paradigm and are constrained by the coverage of training data, making them prone to failure on edge-case configurations after deployment. 
To mitigate such risks, it is essential to expose high-quality failure modes and convert the resulting failures into supervisory data for model enhancement.
Existing studies largely stop at failure detection and lack a mechanism for leveraging discovered failures for model repair. Moreover, prior work possesses limitation for failure exposure, which either rely on manually curated benchmarks with limited coverage, or lack sufficient input effectiveness and diversity for automated testing.

We propose \system{}, the first analysis framework for VLA enhancement, which integrates distance-aware model testing for failure exposure and agent-based data synthesis for model finetunning. 
First, \system{} estimates input uncertainty based on the distance to training samples, and combines uncertainty ranking with redundancy elimination to build compact test sets that expose diverse failures. 
Then, \system{} integrates a VLM-based agent with the capacity of abstraction and planning, to repair the failures into successful executions. 
\system{} abstracts failure trajectories into structured semantic representations, and plans parameterized repair-skill sequences, which are then realized as executable trajectories through inverse kinematics and motion execution. The resulting successful trajectories are automatically labeled and used to fine-tune the original VLA model, yielding an enhanced VLA model. Evaluation on four representative robotic manipulation tasks shows that 92.33\% of the in-simulation test cases generated by \system{} trigger VLA failures, and \system{} improves test coverage over the state-of-the-art tool by 18.93\%. By fine-tuning VLA models with synthetic data derived from failed test cases, \system{} further enhances the overall success rate of VLA models by 49.25\%. When deployed on real hardware, the simulation-enhanced models improve the success rate over the original VLA models by 57.50\%, demonstrating an effective and low-cost direction for VLA enhancement.

\end{abstract}

\begin{IEEEkeywords}
Vision-Language-Action Models, Robotic Manipulation, Test Input Selection, Model Enhancement.
\end{IEEEkeywords}

\section{Introduction}
\label{sec:introduction}

Vision-Language-Action (VLA) models have emerged as one of the key technologies for end-to-end robotic manipulation, which takes visual observations and natural language instructions as input and directly outputs control actions~\cite{ma2024survey}. Like other DNN-based systems, VLA models follow a data-driven paradigm in which their behavior is largely determined by the training data~\cite{guan2025efficient}. Since no dataset can exhaust the diversity of real-world scenes, objects, and instructions, VLA models are prone to failure in edge-case or long-tail configurations after deployment~\cite{wang2025vlatest},
which impedes their reliable use across real-world robotic applications. 
Mitigating such risks therefore requires a low-cost and scalable way to enhance the VLA model, which efficiently obtains high-quality failure modes and turns the failures into supervisory data for model enhancement~\cite{humeniuk2024simulation}.

Despite the severe impact of VLA failures, existing studies~\cite{liu2025eva,wang2025vlatest,zhang2025robustvla,fei2025libero,valle2025evaluating,zeng2025diagnose} largely stop at failure exposure and lack the support for leveraging the discovered failures for model repair. 
For failure exposure, existing studies can be broadly divided into two categories, both of which have notable limitations. The first category relies on manually-curated benchmarks, such as LIBERO-Plus~\cite{fei2025libero}, RobustVLA~\cite{zhang2025robustvla}, Eva-VLA~\cite{liu2025eva}, ViFailback~\cite{zeng2025diagnose}, and VLA-Arena~\cite{zhang2025vla}. These benchmarks require substantial human effort, cover only limited configuration dimensions, and are therefore poorly suited for exposing failures at scale. The second category employs automated testing frameworks, such as VLATest~\cite{wang2025vlatest}, which randomly generate test inputs. However, echoing a long-standing concern in DNN testing, these frameworks pay insufficient attention to the diversity of selected inputs. In particular, many failures stem from the same underlying defect, which causes the resulting test inputs to contain substantial redundancy and reduce testing effectiveness and efficiency.
For the model repair, there is a lack of mechanisms for automatically turning discovered failure trajectories into successful executions that can serve as supervisory data, leaving a gap between failure exposure and model enhancement.

In this paper, we propose \system{}, an analysis framework for VLA enhancement, which integrates distance-aware model testing for failure exposure and agent-based data synthesis for model finetuning. 
With this capability, \system{} enables automatic failure exposure and enhancement of various VLA models before deployment, thereby improving their robustness and mitigating potential risks that may cause physical damage.

Our key observations is that (1) the distance in the model's latent space reveals both the failure-revealing potential and the diversity of test inputs. The distance from a candidate to its nearest training samples reflects how weakly the candidate is supported by the training distribution and thus its likelihood of triggering a failure, while the pairwise distance between candidates indicates whether they may originate from the same underlying defect.
(2) a manipulation failure can be decomposed into a set of localized failure modes (e.g., a grasp failure may arise from object mislocalization, gripper-width mismatch, or end-effector misalignment), each of which can be addressed by a predefined repair primitive. 
Hence, repairing a failure can be reduced to composing a set of primitives rather than hand-crafting a script for each case.

\system{} takes the original VLA model and its training data as input and produces an enhanced model through two phases. In the first phase, \system{} starts from a randomly-generated test candidate pool, estimates each candidate's uncertainty based on its distance to the nearest training samples, refines high-uncertainty candidates through uncertainty sorting, and eliminates redundancy between candidates to obtain a compact test set that effectively exposes diverse failures (see Section~\ref{TestSelection}).
In the second phase, \system{} integrates a VLM-based agent with abstraction and planning capabilities to repair failures into successful executions. 
The agent first abstracts the failure trajectories into structured semantic representations capturing causal relations, state changes, subgoal decomposition, and state abstraction. The agent then composes the parameterized repair skill sequences from a library of perception, motion, and interaction primitives, which are grounded into executable trajectories via inverse kinematics and motion generation.
Successfully-executed trajectories are automatically annotated as supervisory samples and used to fine-tune the original VLA model, producing an enhanced VLA model (see Section~\ref{DataSynthesis}).

\para{Evaluation}
We evaluate \system{} on four representative robotic manipulation tasks. The results show that \system{} effectively reveals diverse failures and can be leveraged to enhance VLA models. On average, 92.33\% of the generated in-simulation test cases cause VLA models to fail the task, and \system{} improves failure coverage over the state-of-the-art testing tool VLATest by 18.93\%. By fine-tuning VLA models with synthetic data derived from failed test cases, \system{} enhances the overall success rate of VLA models by 49.25\%.

Furthermore, \system{} supports a sim-to-real transfer pipeline that reduces reliance on labor-intensive real-world data collection and annotation while lowering the cost of testing on physical robots. To evaluate this capability, we deploy VLA models enhanced solely with simulation-generated data on real-world hardware consistent with the simulation settings. Compared with the original VLA models, the simulation-enhanced models improve the real-world success rate by 57.50\%, which indicates that simulation-based enhancement remains effective in real-world deployment and offers a promising low-cost direction for enhancing VLA models.

In summary, this paper makes the following contributions.

\begin{itemize}
    \item We propose \system{}, the first analysis framework for VLA enhancement, which reveals diverse failures and leverages discovered failures for model enhancement, to the best of our knowledge.

    \item We conduct an extensive evaluation to demonstrate the effectiveness of \system{}. Moreover, sim-to-real transfer evaluation shows that simulation-enhanced VLA models remain effective in real-world scenarios, highlighting a low-cost direction for VLA enhancement.
    \item We will release the artifact and dataset of \system{} to facilitate future research.
\end{itemize}

\section{Preliminary}
\label{sec:Preliminary}

\subsection{VLA Model for Robotic Manipulation}
\label{sec:VLA}

Robot manipulation is a core problem in robotics research~\cite{ma2024survey, xu2024survey,liu2025aligning, wu2025human}, with broad applications in industrial automation~\cite{dzedzickis2021advanced,goel2019robotics}, healthcare~\cite{holland2021service}, and logistics~\cite{dhaliwal2020rise}.
As a class of foundation models for robot manipulation, Vision-Language-Action (VLA) models have rapidly emerged~\cite{xu2024survey}. VLA models enable robots to understand high-level human instructions, reason about spatial relationships, and execute complex manipulation tasks in dynamic environments~\cite{shao2025large}. For instance, a VLA model is expected to follow an instruction such as “\textit{move the cup next to the laptop on the desk}”.
Existing VLA approaches can be broadly divided into two types~\cite{zhang2025generative}, both of which are targeted by \system{} in this work.

\begin{enumerate}
    \item \textbf{Monolithic VLA models}~\cite{kim2024openvla,o2024open}. Visual observations, language instructions, and robot states are jointly fed into a single unified VLA model, such as OpenVLA~\cite{kim2024openvla} or RT-1~\cite{o2024open}, which integrates all modalities and produces executable actions through autoregressive or parallel decoding.

    \item \textbf{Hierarchical VLA models}~\cite{gr00tn1_2025,intelligence2025pi_}. This type of VLA model, such as $\pi_{0.5}$~\cite{intelligence2025pi_}, consists of two components. The first is a planning component, typically instantiated as a high-performance vision-language model (VLM), which performs semantic reasoning and translates a complex task into a sequence of basic actions. The second is a policy module, such as a diffusion model or a monolithic VLA model, which converts these basic actions into executable motor commands to complete the task.
\end{enumerate}

\subsection{Test Input Selection and Model Enhancement}

\para{Test Input Selection}
Test input selection is a widely-adopted technique for improving the efficiency and effectiveness of testing deep neural network (DNN) systems~\cite{attaoui2023black}. Under a constrained testing budget, test input selection aims to identify test inputs with the highest potential to reveal defects in the model under test~\cite{li2024distance}. In essence, this technique distills a large candidate test set into a compact subset whose members exhibit strong failure-revealing capability~\cite{attaoui2025search}.
Formally, given a VLA model under test $F$ and a testing budget $\beta$, the goal is to select a subset $D_S$ from a candidate test set $D^C$ of size $N$ such that:
\begin{equation}
\footnotesize
D_S \subseteq D^C, \quad |D_S| = \beta < N
\end{equation}
while maximizing the capability of detecting defects in $F$.

\para{Model Repair and Enhancement}
The ultimate purpose of testing VLA models goes beyond failure detection and extends to the behavioral improvement of the model under test~\cite{humeniuk2024simulation}. Following prior DNN repair studies~\cite{li2022hybridrepair, attaoui2025search}, we focus on model repair that corrects a model's erroneous behaviors through data-augmentation-based parameter updates while preserving its network architecture. Within this view, every discovered failure trajectory can be used to generate supervisory data to retrain or fine-tune the model for performance enhancement. 
Intuitively, retraining or fine-tuning achieves greater improvement when the supplementary data spans a broader spectrum of failure modes~\cite{attaoui2023black}. Therefore, test input selection is also valuable in the repair and enhancement process. Specifically, by prioritizing data with high failure-revealing potential, effective model improvement can be achieved with reduced labeling and computational costs~\cite{li2022hybridrepair}.

\begin{figure*}[t]
\centering
\includegraphics[width=7.2 in]{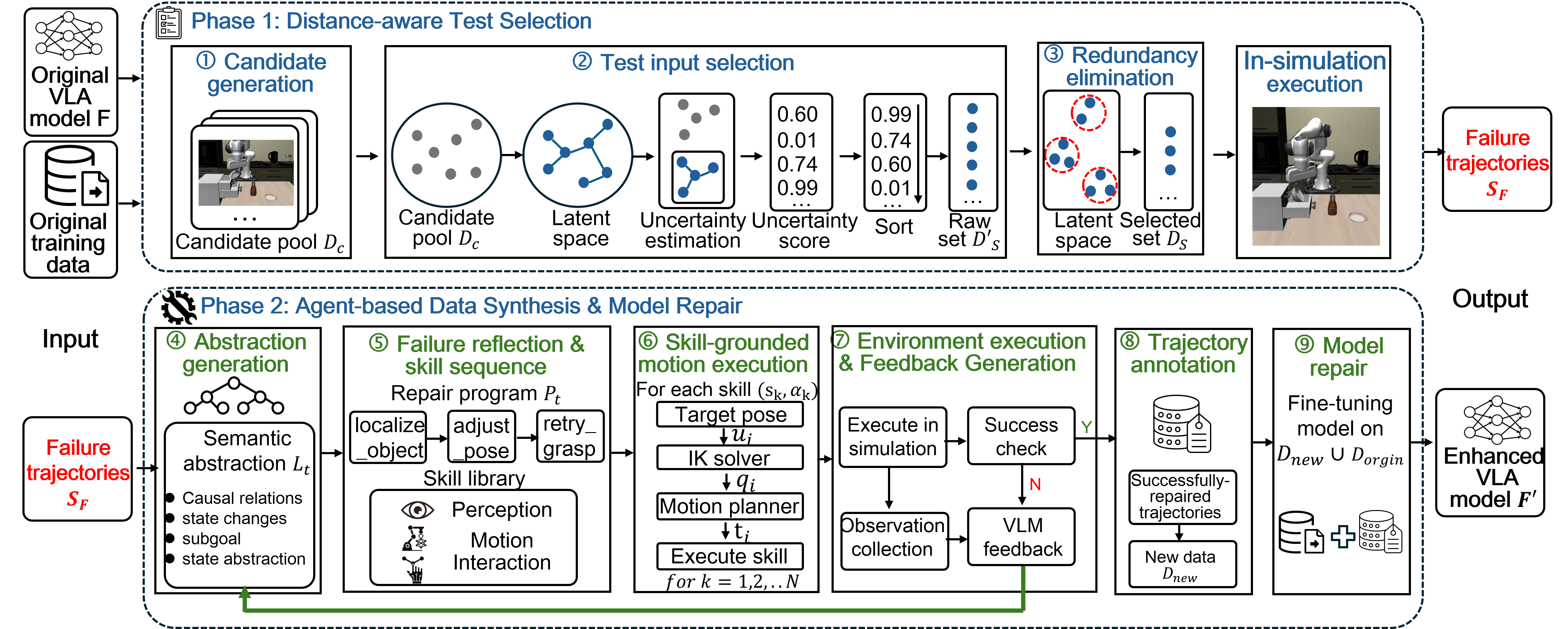}
\caption{The workflows of \system{}} 
\label{fig:overview}
\vspace{-2mm}
\end{figure*}

\section{Design of \system{}}
\label{sec:insight}

\subsection{Key Observations}
\label{sec:observation}

\para{Observation 1: For failure exposure, latent space distance reveals both the failure-revealing potential and the diversity of test candidates}
Since existing studies (e.g., VLATest~\cite{wang2025vlatest}) generate test inputs at random and ignore their failure-revealing capability, the effectiveness of the generated test inputs is limited.
Similar to traditional DNN testing, multiple test inputs may originate from the same underlying defect~\cite{li2024distance}. Therefore, randomly-generated test inputs can contain substantial redundancy and expose only a fraction of distinct failures.
Effective in-simulation testing therefore requires a selection strategy that simultaneously prioritizes failure-revealing test inputs and reduces redundancy among them.

We observe that the distances computed in the model's latent space support both requirements. \textit{(1) Candidate-to-training distance.} A candidate far from its nearest training samples lies in a region where the model has received limited supervision and is therefore more likely to expose a distinct failure~\cite{li2024distance, cohen2020detecting}. This distance reflects the uncertainty of test inputs.
Following prior work~\cite{li2024distance}, \system{} uses this distance as a failure-revealing score.
\textit{(2) Candidate-to-candidate distance.} Two candidates that lie closed to each other in the latent space share similar model representation vectors and tend to trigger the same underlying defect.
Hence, \system{} utilizes this distance as a direct indicator of redundancy.

\para{Observation 2: A failure trajectory can be extracted into semantic and compositional information that an agent can use for repair}
Using the revealed failures to improve the model itself faces the following challenges.
First, a failure trajectory is a raw, low-level state-action trace that neither directly reflects the root cause nor identifies which aspect of execution has gone wrong. Directly predicting corrective actions from such traces would force the repair module to absorb the full complexity of continuous control~\cite{guan2025efficient}.
Second, hand-crafting repair scripts for individual failure cases does not scale to the diversity of robot manipulation scenarios.

We observe that a failure trajectory carries the following two types of information. \textit{(1) Semantic information.} A failure can be summarized as a set of language-level descriptors, including causal relations, state changes, subgoal decomposition, and state abstraction, which together identify the gap between the intended outcome and the failed execution of the VLA model.
This information can guide an agent in reasoning about and repairing the failure~\cite{wang2025element, sarch2024vlm}.
\textit{(2) Compositional information.} Manipulation failures encountered in practice can be decomposed into a bounded number of localized failure patterns at the perception, motion, and interaction levels, each of which can be addressed by a parameterized repair primitive.
A repair program can therefore be reduced to selecting, ordering, and parameterizing a few such primitives from a fixed skill library, rather than synthesizing actions from scratch.

\subsection{Overview of \system{}}
\label{sec:overview}

\system{} is designed as an in-simulation framework for testing and repairing VLA models in robotic manipulation. As illustrated in Fig.~\ref{fig:overview}, \system{} takes the original VLA model $F$ together with its associated training dataset $D_{\mathrm{orig}}$ as input, and ultimately produces an enhanced model $F'$ with improved robustness against the identified failure modes. To achieve this goal, \system{} operates through two complementary phases.

\para{Distance-aware Test Selection}
In the first phase, \system{} aims to efficiently identify a compact and diverse set of failure-inducing test cases.
Specifically, \system{} first constructs a candidate test pool $D^C$ by leveraging an automated testing tool, namely VLATest~\cite{wang2025vlatest} (see \textcircled{1}).
\system{} estimates the uncertainty score for each candidate in $D^C$ by computing its distance to the nearest training samples in the latent space.
The rationale is that candidates farther from the training distribution are more likely to expose model failures because the model has received limited supervision in those regions~\cite{li2024distance}.
Candidates are ranked by their uncertainty scores in descending order, and the top-ranked inputs are selected to form a preliminary subset $D_S'$ (see \textcircled{2}).
To further improve the diversity of the selected set, \system{} applies a redundancy elimination procedure that identifies and removes test inputs residing in closed proximity within the latent space, as such inputs are likely to trigger the same underlying defect (see \textcircled{3}).
The resulting refined subset $D_S$ is then executed in the simulation environment to obtain the set of failure cases $S_F$.

\para{Agent-based Data Synthesis and Model Repair}
In the second phase, \system{} utilizes the discovered failures to generate synthetic supervisory data through an agent-driven repair process and subsequently uses the synthesized data to fine-tune the model for enhancement.
For each failure case, \system{} replays the failure trajectory in simulation and generates a structured semantic abstraction $L_t$ that captures the causal relations, state changes, subgoal decomposition, and state abstraction (see \textcircled{4}).
Guided by this abstraction, a VLM-based repair planner produces a parameterized repair skill sequence $P_t$, where each skill is drawn from a predefined library of manipulation primitives covering perception, motion, and interaction repair (see \textcircled{5}).
Each skill in $P_t$ is subsequently grounded into executable robot motions through target pose computation, inverse kinematics solving, and motion planning (see \textcircled{6}).
After execution in the simulation environment, if the repair succeeds, the resulting trajectory is automatically annotated as supervised training data.
Otherwise, \system{} uses a VLM to analyze the observations from the repair failure, generate diagnostic feedback, and feed it back to the abstraction and planning modules, forming an iterative repair loop (see \textcircled{7}).
The successfully-repaired trajectories are aggregated into a new dataset $D_{\mathrm{new}}$ (see \textcircled{8}), which is merged with $D_{\mathrm{orig}}$ and used to fine-tune the original model $F$, yielding the enhanced model $F'$ with improved robustness and generalization (see \textcircled{9}).

\section{Approach Detail}
\label{sec:Design}

\subsection{Distance-aware Test Selection}
\label{TestSelection}

\para{Preprocessing and Test Candidate Generation}
\label{CandidateGeneraton}
As a preprocessing step, \system{} uses VLATest\cite{wang2025vlatest}, a state of the art test analysis tool, to generate an initial pool of candidate tests for subsequent selection. Compared with manually-designed benchmarks such as LiBERO-Plus\cite{fei2025libero}, RobustVLA\cite{zhang2025robustvla}, and Eva-VLA\cite{liu2025eva}, VLATest automatically generates many randomized tests over a broader set of scene factors. Specifically, VLATest takes the target VLA model as input and stochastically samples testing configurations to produce candidate test cases.

Let $D^c=\{x_i^c\}_{i=1}^{N^c}$ denote the candidate test set generated in preprocessing, and let $F$ denote the target VLA model. The goal is to select a preliminary subset $D'_s \subset D^c$ whose inputs are more likely to expose failures in $F$. Our key insight is that the distance from a test input to its nearest training sample can serve as a proxy for model uncertainty, because a larger distance often indicates weaker coverage by the training data. Intuitively, inputs far from the training data are more likely to lie in underrepresented regions and thus to trigger erroneous behavior. Accordingly, we assign each candidate $x_i^c$ an uncertainty score defined by its distance to the nearest training sample, and rank all candidates in descending order. The highest ranked candidates are selected to form $D'_s$.

\para{Latent Representation}
To compute distance-based uncertainty scores, we first need an appropriate latent feature space. Prior studies show that the choice of latent space strongly affects the effectiveness of distance metrics\cite{frosst2019analyzing}. Unlike conventional DNNs, which often use the penultimate layer as the latent space\cite{frosst2019analyzing,jiang2018trust}, VLAs fuse visual observations and language instructions before passing the fused representation to an action head. Therefore, \system{} uses the \textit{fused hidden state} immediately before the action head as its latent feature space. This representation encodes the final integration of visual and linguistic information before action prediction, capturing the model's decision relevant internal state.

Let $f$ denote the composite function of all VLA layers before the action head, whose output is the fused hidden state. For any training or test input $x_i$, its latent feature $z_i$ is defined as:
\begin{equation}
  \footnotesize
  z_i = f(x_i)
\end{equation}

\para{Distance Metric}
We use cosine distance in this latent space because VLA embeddings encode fused multimodal semantics, for which direction is often more informative than vector magnitude. This angular metric is insensitive to norm variations caused by input scale or modality specific encoding, providing a more stable measure of semantic dissimilarity. Formally, the distance between two inputs $x_i$ and $x_j$ is defined as:
\begin{equation}
  \footnotesize
  d(x_i, x_j) = 1 - \frac{z_i \cdot z_j}{\|z_i\|\|z_j\|}
\end{equation}

\para{Training Support Degree}
With the distance metric defined, we quantify how strongly the training data supports the VLA prediction for a given test input. The rationale is that a test input closer to training samples in the latent space is more strongly supported by the training distribution~\cite{cohen2020detecting, papernot2018deep, romano2006exploring, frosst2019analyzing}. Following NED~\cite{karpusha2020calibrated}, we estimate this support using normalized exponential weights over the distances between the test input and its nearest training samples. Since VLA outputs include both discrete action types (e.g., OpenVLA, RT-1) and continuous action values (e.g., $\pi_{0.5}$), we define the training support degree for each case separately.

For discrete action types, given a candidate input $x_i^c$, let $\mathcal{N}_T(x_i^c)$ denote its $K_T$ nearest neighbors in the training set. The weighted nearest neighbor support probability for action class $c$ is defined as:
\begin{equation}
\footnotesize
p_c^*(x_i^c) =
\frac{
\sum_{x_j^T \in \mathcal{N}_T(x_i^c)}
\exp(-d(x_i^c,x_j^T)/\tau)\,\mathbb{I}(a_j^T=c)
}{
\sum_{x_j^T \in \mathcal{N}_T(x_i^c)}
\exp(-d(x_i^c,x_j^T)/\tau)
}
\end{equation}
where $d(x_i^c, x_j^T)$ denotes the cosine distance in the latent space, $\tau$ controls the decay rate of the distance based weights, and $\mathbb{I}(\cdot)$ is the indicator function. If the VLA predicts action class $\hat{c}_i^c$ for $x_i^c$, its discrete action support degree is given by $S(\hat{c}_i^c)=p_{\hat{c}_i^c}^*(x_i^c)$.

For continuous action values, we define the training support degree as:
\begin{equation}
\footnotesize
S(\hat{a}_i^c) =
\frac{
\sum_{x_j^T \in \mathcal{N}_T(x_i^c)}
\exp(-d(x_i^c,x_j^T)/\tau)
\exp(-\|\hat{a}_i^c - a_j^T\|^2/\gamma)
}{
\sum_{x_j^T \in \mathcal{N}_T(x_i^c)}
\exp(-d(x_i^c,x_j^T)/\tau)
}
\end{equation}
where $\hat{a}_i^c=F(o_i^c,l_i^c)$ is the action predicted from observation $o_i^c$ and instruction $l_i^c$, $a_j^T$ is the expert action of the neighboring training sample, and $\gamma$ controls the scale of the action similarity kernel.

\para{Uncertainty Score}
Based on the training support above, we define the uncertainty score for each test input. Since VLAs may output either discrete or continuous actions, we define the score for each case separately.

For discrete action outputs, let $c'$ denote the non-predicted action class with the highest support. We define uncertainty as the ratio between the support of $c'$ and the support of the predicted class $\hat{c}_i^c$:
\begin{equation}
\footnotesize
U(x_i^c) =
\frac{p_{c'}^*(x_i^c)}
{p_{\hat{c}_i^c}^*(x_i^c)}
\end{equation}
The intuition is that if a non-predicted class receives support comparable to or higher than the predicted class, the prediction is likely less reliable.

For continuous action outputs,  the ratio-based formulation does not directly apply. Instead, we define uncertainty as the negative logarithm of the predicted action support:
\begin{equation}
\footnotesize
U(x_i^c) =
-\log(S(\hat{a}_i^c) + \epsilon)
\end{equation}
where $\epsilon$ is a small constant used to avoid taking the logarithm of zero.

In both cases, a higher uncertainty score indicates weaker training support and a higher likelihood of prediction error. Therefore, test inputs with higher uncertainty are prioritized for selection.

\para{Redundancy-aware
Diversity Score}
Since multiple inputs in $D'_s$ may expose the same underlying defect, we quantify redundancy to reduce duplicate selections. Inputs closed in the latent space tend to share similar properties and may trigger the same type of failure. For each $x_i \in D'_s$, we compute a redundancy-aware
diversity score as:
\begin{equation}
\footnotesize
R(x_i) =
\sum_{x_j \in \mathcal{N}_R(x_i)}
d(x_i, x_j)
\end{equation}
where $\mathcal{N}_R(x_i)$ denotes the $K_R$ nearest neighbors of $x_i$ within $D'_s$, and $d(\cdot,\cdot)$ denotes cosine distance in the fused hidden state space.

A larger $R(x_i)$ indicates that $x_i$ differs more from its neighbors and is more likely to reveal a distinct failure type, so it should be retained. Conversely, a smaller $R(x_i)$ suggests potential redundancy, so removing it is less likely to reduce failure diversity.

\begin{algorithm}[t]
\footnotesize
\caption{Test input selection algorithm in \system{} }
\label{alg:datis}
\begin{algorithmic}[1]
\Require VLA $F$; training set $\{(x_i^T, a_i^T)\}_{i=1}^{N_T}$;
candidate test set $D^C$; testing budget $\beta$;
hyper-parameters $\tau, K_T, K_R$;
for discrete actions: action class set $\mathcal{C}$;
for continuous actions: $\gamma$, $\epsilon$
\Ensure Selected test input set $D_S$

\State $U \gets \emptyset$

\ForAll{$x_i^C \in D^C$}
    \State $z_i^C \gets f(x_i^C)$
    \Comment{Distance computation}
    \State Find $K_T$ nearest neighbors $\mathcal{N}_T(x_i^C)$
    via $d(\cdot,\cdot)$
    \State $w_j \gets \exp(-d(x_i^C, x_j^T)/\tau)$,
    $\forall x_j^T \in \mathcal{N}_T(x_i^C)$

    \If{action type is discrete} \Comment{Uncetainty for Discrete actions}
        \State $\hat{c}_i \gets F(x_i^C)$
        
        \ForAll{$c \in \mathcal{C}$}
            \State {\footnotesize
            $\displaystyle p_c^*(x_i^C) \gets
            \frac{\sum_j w_j \,
            \mathbb{I}(a_j^T\!=\!c)}
            {\sum_j w_j}$}
        \EndFor
        \State $c' \gets
        \arg\max_{c \neq \hat{c}_i} p_c^*(x_i^C)$
        \State $U(x_i^C) \gets
        {p_{c'}^*(x_i^C)}\,/\,{p_{\hat{c}_i}^*(x_i^C)}$

    \Else
        \Comment{Uncetainty for Continuous action}
        \State $\hat{a}_i^C \gets F(x_i^C)$
        \State {\footnotesize
        $\displaystyle S(\hat{a}_i^C) \gets
        \frac{\sum_j w_j
        \exp(-\|\hat{a}_i^C \!-\! a_j^T\|^2/\gamma)}
        {\sum_j w_j}$}
        \State $U(x_i^C) \gets
        -\log(S(\hat{a}_i^C) + \epsilon)$
    \EndIf

    \State $U \gets U \cup \{(x_i^C, U(x_i^C))\}$
\EndFor

\State $D_S' \gets \textsc{Select}(D^C, U, \beta)$
\Comment{High-uncertainty selection}

\State $R \gets \emptyset$

\ForAll{$x_i \in D_S'$}
    \State Find $K_R$ nearest neighbors
    $\mathcal{N}_R(x_i)$ in $D_S'$
    \State {\footnotesize
    $\displaystyle R(x_i) \gets
    \sum_{x_j \in \mathcal{N}_R(x_i)} d(x_i, x_j)$}
    \State $R \gets R \cup \{(x_i, R(x_i))\}$
\EndFor

\State $D_S \gets \textsc{Refine}(D_S', R, \beta)$
\Comment{Redundancy elimination}

\State \Return $D_S$
\end{algorithmic}
\end{algorithm}

\para{Test Selection}
\system{} selects final test inputs from the candidate set through two steps.

\textit{Step 1: Uncertainty-based ranking.}
Given a budget $\beta$, we rank candidates in $D^c$ in descending order of uncertainty and select the top $\beta$ candidates to form a preliminary subset $D'_s$.

\textit{Step 2: Redundancy elimination.}
We then filter $D'_s$ by removing inputs that are closed in the fused hidden state space and are likely to expose the same underlying defects. The filtering strength is adjusted according to the testing budget $\beta$ to balance the number of retained inputs and failure diversity.

Algorithm~\ref{alg:datis} presents the test input selection procedure of \system{}. The algorithm takes a candidate test set $D^c$ and a target VLA model $F$ as input, and outputs the selected test set $D_s$. For each candidate $x_i^c \in D^c$, the algorithm extracts its fused hidden state $z_i^c=f(x_i^c)$ and finds its $K_T$ nearest training neighbors $\mathcal{N}_T(x_i^c)$ using cosine distance. It then computes an exponential distance weight $w_j$ for each neighbor and computes uncertainty according to the action type.

For discrete actions, the algorithm obtains the predicted class $\hat{c}_i^c$, computes the weighted support $p_c^*$ for each class, and defines uncertainty as the ratio between the highest non predicted class support and the predicted class support. For continuous actions, the algorithm obtains the predicted action $\hat{a}_i^c$, computes the kernel weighted support $S(\hat{a}_i^c)$, and sets the uncertainty to $-\log(S(\hat{a}_i^c)+\epsilon)$. In both cases, the resulting score is added to the list $U$.

After all candidates are scored, the algorithm ranks them by uncertainty and selects the top $\beta$ candidates to form $D'_s$. To reduce redundancy, for each $x_i \in D'_s$, the algorithm finds its $K_R$ nearest neighbors within $D'_s$ and computes the sum of distances to them as the inverse redundancy score $R(x_i)$. Finally, the algorithm filters $D'_s$ using $R(x_i)$ under the testing budget $\beta$, and returns the resulting set as the final selected test set $D_s$.
If redundancy elimination produces fewer test inputs than buget $\beta$, \system{} fills the remaining slots from the unselected test cases from candidate pool $D_C$ according to uncertainty score, until $|D_{s}| = \beta$.

\subsection{Agent-based Data Synthesis and Model Repair}
\label{DataSynthesis}

\para{Abstraction Generation}
A failure trajectory collected during simulation testing is a raw execution trace that records low-level state action pairs without explicit semantic annotations. Such traces provide limited guidance for downstream repair because they do not explicitly identify failure relevant factors or indicate which aspects of execution deviate from the intended behavior. Therefore, a structured semantic representation is needed to convert low-level signals into a form that supports systematic repair reasoning.

To obtain this representation, \system{} replays the failure trajectory in the simulator and restores the execution state at each timestep. \system{} then uses ICAL~\cite{sarch2024vlm} to derive a high-level semantic abstraction of the failure. Specifically, the abstraction function takes the current observation $o_t$, task instruction $l$, and execution history $\tau_{\text{history}}$ as input, and produces:
\begin{equation}
\footnotesize
L_t = \mathcal{A}(o_t, l, \tau_{\text{history}})
\end{equation}

The resulting abstraction $L_t$ contains four components: causal relations, state changes, subgoal decomposition, and state abstraction. Together, they describe the task structure and intended outcome, failure-relevant factors, key transitions in the environment and robot state, intermediate objectives, and state attributes related to the failure.

Conceptually, $L_t$ serves as a semantic intermediate representation between low-level execution failures and high-level repair decisions. By distilling raw traces into structured and interpretable semantic information, $L_t$ allows the repair module to reason about the failure without directly processing opaque state action sequences. The generated $L_t$ is used as the conditional input to the next stage, where it guides repair skill selection and parameterization to construct an executable repair program.

\para{Failure Reflection and Skill Sequencing}
Given the structured abstraction $L_t$, \system{} determines how to repair the corresponding failure. Instead of directly predicting low-level robot actions, \system{} formulates repair as constructing an executable program in a predefined skill space, avoiding the need for the repair module to directly handle continuous control. This decomposition separates high-level repair reasoning from low-level motion generation, allowing them to be handled separately.

The skill space is defined by a predefined repair skill library, which contains 16 parameterized repair primitives organized into three categories according to the manipulation aspect they address. \textit{Perception Repair} skills address errors in robot state estimation, including re-estimating object and target poses, refining grasp contact locations, and identifying valid placement or insertion regions. \textit{Motion Repair} skills correct errors in end effector trajectories and orientations, including moving to pre-grasp or pre-placement poses, adjusting the end effector pose, changing the approach direction, and aligning the end effector with the object or target axis. \textit{Interaction Repair} skills address failures arising from physical contact between the robot and objects, including retrying grasps, adjusting gripper width, stabilizing grasps, placing with offsets, releasing objects, and verifying task success.

Each skill $s_i \in \mathcal{S}$ takes typed arguments $\alpha_i$ (e.g., object identifiers, spatial offsets, or gripper parameters), and is associated with one or more manipulation primitives, as detailed in Table~\ref{tab:skill_library}.

A repair program is defined as an ordered sequence of parameterized skill invocations:
\begin{equation}
\footnotesize
P_t = \big((s_1,\alpha_1), \dots, (s_k,\alpha_k)\big)
\end{equation}
where each pair $(s_i,\alpha_i)$ specifies a repair skill and its arguments.

\system{} generates the repair program based on the failure abstraction from the preceding stage. Specifically, a VLM repair planner takes $L_t$, the current observation $o_t$, and the task instruction $l$ as input, and outputs the skill sequence:
\begin{equation}
\footnotesize
P_t = \mathcal{G}(L_t, o_t, l)
\end{equation}
Here, $L_t$ provides the planner with failure-relevant factors, relevant state variables, and subgoal structure, thereby constraining skill selection, ordering, and parameterization. The resulting $P_t$ is a compositional repair program in which each skill addresses a localized aspect of the failure, and the sequence forms a coherent multi-step repair strategy. Specifically, $P_t$ is required to emit anordered list of skill invocations, where each entry conforms to the schema. 
\begin{center}
\footnotesize
\texttt{\{ "step": $k$, "skill": $s_k \in \mathcal{S}$, "arguments": $\alpha_k$ \}}
\end{center}

\system{} generates structured prompts to guide VLM to produce the repair program $P_t$. The prompts are organized as the three-slots template $(L_t, o_t, l)$, and released in our open-source repository.

The generated repair program $P_t$ is then passed to the next stage, where each skill invocation is grounded into physically-executable robot actions.

\begin{table*}[t]
\centering
\caption{Compact Repair Skill Library}
\label{tab:skill_library}
\resizebox{\linewidth}{!}{
\begin{tabular}{lllll}
\toprule
\textbf{Category} & \textbf{Skill} & \textbf{Description} & \textbf{Arguments} & \textbf{Covered Primitives} \\
\midrule

\multirow{4}{*}{Perception Repair}
& relocalize\_object(A) 
& Re-estimate the pose of object A 
& object A 
& All \\

& relocalize\_target(B) 
& Re-estimate the pose of target B 
& target B 
& Move-to, Put-on, Put-into \\

& recenter\_grasp\_point(A) 
& Refine grasp point on object A 
& object A 
& Pick-up \\

& estimate\_target\_region(B) 
& Estimate valid placement/insertion region on B 
& target B 
& Put-on, Put-into \\

\midrule

\multirow{6}{*}{Motion Repair}
& Move-to\_to\_pregrasp(A) 
& Move-to end-effector to pre-grasp pose 
& object A 
& Pick-up \\

& Move-to\_to\_preplace(B) 
& Move-to end-effector to pre-placement pose 
& target B 
& Move-to, Put-on, Put-into \\

& adjust\_end\_effector\_pose 
& Fine-tune pose using offsets 
& dx, dy, dz, droll, dpitch, dyaw 
& All \\

& change\_approach\_direction(A/B) 
& Modify approach direction 
& object A or target B 
& All \\

& align\_with\_object\_axis(A) 
& Align with object A axis 
& object A 
& Pick-up \\

& align\_with\_target\_axis(B) 
& Align with target orientation 
& target B 
& Move-to, Put-on, Put-into \\

\midrule

\multirow{6}{*}{Interaction Repair}
& retry\_grasp(A) 
& Re-execute grasp action 
& object A 
& Pick-up \\

& change\_gripper\_width(w) 
& Adjust gripper width 
& width w 
& Pick-up \\

& stabilize\_grasp(A) 
& Stabilize grasped object 
& object A 
& Pick-up, Move-to \\

& place\_with\_offset(B) 
& Adjust placement position 
& target B, dx, dy, dz 
& Put-on, Put-into \\

& release\_object() 
& Release object from gripper 
& None 
& Put-on, Put-into \\

& verify\_success(A, B, type) 
& Check task success 
& object A, target B, type 
& All \\

\bottomrule
\end{tabular}
}
\end{table*}

\para{Skill Grounded Motion Execution}
Once the repair program $P_t$ has been constructed, \system{} translates each symbolic skill invocation into physically-executable robot actions. Because each library skill has a predefined deterministic execution procedure, this translation follows a fixed pipeline driven by skill parameters, rather than relying on additional learned policy inference.

For each skill and argument pair $(s_i,\alpha_i)$ in $P_t$,  \system{} performs a grounding process with three stages. First,  \system{} computes a target end effector pose $u_i$ from the skill semantics and its arguments:
\begin{equation}
\footnotesize
u_i = f_m(s_i, \alpha_i)
\end{equation}
where $f_m$ maps the skill type and parameters to a goal pose in Cartesian space.

Second, an inverse kinematics solver converts the target pose into a feasible configuration $q_i$:
\begin{equation}
\footnotesize
q_i = \mathrm{IK}(u_i)
\end{equation}

Third, a motion generator produces a collision-free trajectory $\tau_i$ from the configurations:
\begin{equation}
\footnotesize
\tau_i = \mathrm{Motion}(q_{i-1}, q_i)
\end{equation}

This layered architecture separates symbolic repair reasoning from physical execution. The reflection layer generates the skill sequence at the symbolic level, specifying which repair actions to perform and which parameters to use. The execution layer uses the inverse kinematics solver and motion generator to realize these commands physically, ensuring that each commanded pose is kinematically reachable and that the connecting trajectory is collision-free. Since all skills are predefined primitives with deterministic grounding logic, the translation from symbolic program to physical execution does not require additional learned policy inference. 
Note that the skill library is designed to be generic and backward-compatible, please refer to the discussion~\ref{sec:discuss}.

\para{Environment Execution and Feedback Generation}
After obtaining the grounded trajectories, \system{} executes each trajectory in the simulator and updates the robot state and environmental observations after each skill invocation.

To determine whether the repair succeeds, \system{} uses the task-specific success checkers defined in VLATest\cite{wang2025vlatest}. Each task has a predefined completion criterion that evaluates objective environmental conditions, such as whether the target object reaches the designated position or satisfies the required spatial constraints. The success checker takes the current environment state as input and returns whether the task has been completed successfully. Because the judgment relies on environment groundtruth rather than model-generated assessments, it improves reliability.

If the checker returns success, the complete execution trajectory is forwarded to the next stage for training data construction. Otherwise, \system{} invokes a VLM to analyze the observation of current execution and generate structured diagnostic feedback. Specifically, the VLM takes the current observation, task instruction, and execution history as input, and produces two types of feedback. The \textit{error signal} describes the failure type and its semantic failure information, such as grasp slippage, placement misalignment, or collision. The \textit{state deviation} characterizes the discrepancy between the current state and the expected goal state over key variables, such as object pose, end effector position, and target region alignment.

The observation and VLM feedback are then sent back to the abstraction generation module, initiating a new round of failure analysis and repair planning. This closed loop mechanism enables \system{} to iteratively refine its repair strategy until the success condition is met or the predefined maximum number of attempts is reached.

\para{Automatic Trajectory Annotation}
This stage is triggered only when the success checker confirms that the repair has achieved the task objective. For each successful execution, \system{} automatically converts the recorded trajectory into supervised training data $D_{\text{new}}$.

The resulting dataset has two advantage. First, annotation is automatic because labels are derived directly from simulation execution traces without manual labeling. Second, the data is failure-aware. Since successful trajectories originate from the repair process, they cover scenarios where the model previously failed and provide targeted supervision for these failure cases. 
The constructed dataset $D_{\text{new}}$ is then passed to the next stage for fine tuning.

\para{Model Enhancement}
The final stage of \system{} uses the automatically-annotated trajectories to fine-tune the VLA model. The newly-generated dataset $D_{\text{new}}$ is merged with the original training data $D_{\text{orig}}$ to form a combined fine-tuning corpus. Retaining the original data helps mitigate catastrophic forgetting of previously-acquired capabilities, while the new samples provide targeted supervision for the identified failure cases. 

Since $D_{\text{new}}$ is derived from failure repair, this fine-tuning process provides three complementary benefits. First, the new samples provide correct demonstrations for scenarios in which the model previously failed, helping reduce the corresponding errors. Second, it improves robustness by exposing the model to diverse failure modes encountered during repair. Third, repair trajectories collected across various tasks and environmental configurations expand the training distribution and can improve generalization.

\section{Evaluation}
\label{sec:evaluation}

\subsection{Experiment Setup}
\label{sec:setup}

\para{Research Questions}
We formulate the following research questions (RQs) for the evaluation.

\begin{itemize}
    \item RQ1. How effective is \system{} in revealing VLA failures?
    \item RQ2. How effective is \system{} in repairing VLA failures?
    \item RQ3. How effective is the sim-to-real transfer of in-simulation optimized VLA models in the real world?
\end{itemize}

\para{Robotic manipulation tasks}
Similar to prior work~\cite{wang2025vlatest,zhang2025robustvla,fei2025libero}, we evaluate \system{} on representative robotic manipulation tasks, including (1) \textit{Pick up an object}, (2) \textit{Move object A to object B}, (3) \textit{Put object A on top of object B}, and (4) \textit{Put object A into object B}. Nevertheless, \system{} is designed as a general framework that can incorporate other complex tasks with only lightweight modifications.

\para{Target VLA Models}
To answer the research questions, we test a diverse set of open-source VLA models, including OpenVLA~\cite{kim2024openvla}, RT-1~\cite{o2024open}, and $\pi_{0.5}$~\cite{intelligence2025pi_}. These models cover different dimensions, including model architecture, model size, and reasoning capacity. We choose these VLA models for evaluation because they are widely regarded as representative monolithic or hierarchical models~\cite{guan2025efficient, zhang2025robustvla,fei2025libero}.

\para{Implementation of \system{}}
We implemented \system{} in Python 3.8.10 with approximately 9,600 lines of code. All simulation-based experiments, corresponding to RQ1 and RQ2, were conducted on an Ubuntu 20.04 server equipped with an Intel i9-10980XE CPU at 3.0 GHz, an RTX 3090 GPU, and 250 GB of RAM. Hardware experiments, corresponding to RQ3, were conducted on a Franka robotic arm.
Given the high computational cost, we independently executed each evaluation at least five times. 
We use GPT-5-mini as the backbone VLM for agent-based data synthesis and model repair. 

\para{Evaluation metric} 
(1) \textit{Failure rate} is used as the primary metric to measure the failure-revealing capability of test inputs. In our evaluation, the failure rate is defined as the proportion of test cases that cause the target VLA model to fail the task in execution. 
Note that a higher failure rate indicates stronger failure-revealing effectiveness rather than poorer performance of tested VLA model.
(2) \textit{Trajectory coverage} is used as the metric to measure the diversity of test inputs. 
The trajectory coverage metric is caculated by the method proposed by VLATest~\cite{wang2025vlatest}.
(3) \textit{Success rate improvement} is used as the metric to measure the model enhancement benefited from model repair. The success rate is defined as the proportion of test cases that cause the target VLA model to pass the task in execution.

\subsection{RQ1: Effectiveness in Revealing VLA failures}
\label{sec:RQ1}

\para{Overall Effectiveness}
For each manipulation task, we run \system{} on three VLA models with a testing budget of 100 test cases for each task-model pair, resulting in a total of 1,200 evaluations. This evaluation budget is similar to those adopted in state-of-the-art studies~\cite{wang2025vlatest,fei2025libero,zhang2025robustvla, humeniuk2024simulation}.
We then calculate the total number of failures and the number of failures across different robotic manipulation tasks.

\begin{table}[t]
\centering
\footnotesize
\caption{The performance of \system{} across different VLA models, for evaluating the overall effectiveness in revealing VLA failures}
\label{tab:overall}
\setlength{\tabcolsep}{3.5pt}

\begin{tabular}{lcccccccc}
\toprule
\textbf{VLA Models} & \multicolumn{3}{c}{\textbf{All Tests}} & \multicolumn{4}{c}{\textbf{Failures in Different Tasks}} \\
\cmidrule(lr){2-4}\cmidrule(lr){5-8}
                   & Pass   & Fail   &Total    & Pick up   & Move to   & Put on  & Put in\\
\midrule
RT-1-X             & 38     & 362     & 400    & 85        & 86        & 100      & 91       \\
OpenVLA-7B         & 39     & 361     & 400    & 91        & 82        & 100      & 88      \\
$\pi_{0.5}$        & 15     & 385     & 400    & 93        & 97        & 99       & 96      \\
 \hline
\textbf{Total}     &92      & 1108    & 1200   & 269       & 265       & 299      & 275       \\ 
\bottomrule
\end{tabular}

\vspace{-2mm}
\end{table}

Table~\ref{tab:overall} presents the effectiveness of \system{} in revealing failures across different VLA models.
Overall, \system{} effectively produces failure-revealing test inputs, with 92.33\% (1108/1200) of the generated test cases resulting in VLA failures.

When applied to different VLA models, \system{} maintains strong performance.
\system{} is effective on monolithic models such as RT-1 and OpenVLA, where 90.50\% (362/400) and 90.25\% (361/400) of the test cases result in VLA failures, respectively. The failure rate is higher on the hierarchical model $\pi_{0.5}$, reaching 96.25\% (385/400). These findings show that \system{} exhibits strong generalizability across diverse VLA models, regardless of model architecture, reasoning capability, or model size.

When applied to different tasks, \system{} achieves consistently high failure rates, all exceeding 88\%. Specifically, the failure rates for \textit{Pick-up} and \textit{Move-to} are 89.67\% (269/300) and 88.33\% (265/300), respectively. Higher failure rates are observed for the more complex \textit{Put-on} and \textit{Put-into} tasks, reaching 99.67\% (299/300) and 91.67\% (275/300), respectively. These results indicate that \system{} remains effective across manipulation tasks with different difficulty levels.

Furthermore, we manually inspected all ineffective test cases across the four tasks. The results reveal that most of them, 63.04\% (58/92), are caused by the limited capability of the candidate generation tool, VLATest, integrated into \system{}.
For instance, the candidate generation tool produces a high proportion of ineffective test cases. Under a fixed testing budget, although \system{} filters such cases as aggressively as possible, it cannot fully avoid their inclusion.
This limitation can be mitigated by integrating a more advanced candidate generation tool in future work.

\begin{table}[t]
\centering
\footnotesize
\caption{The comparison results between \system{} and VLATest~\cite{wang2025vlatest} in failure detection and trajectory coverage }
\label{tab:comparison}
\setlength{\tabcolsep}{12pt}

\begin{tabular}{lcc}
\toprule
\textbf{Approach} 
                   & \textbf{Failure detection}   & \textbf{Trajectory coverage}   \\
\midrule
VLATest~\cite{wang2025vlatest}             &  83.62\% (669/800)     & 63.75\%                      \\
\system{}         &90.38\% (723/800)       & 82.68\%                     \\ 
\bottomrule
\end{tabular}

\end{table}

\para{Comparison with Prior Work}
First, we provide a qualitative discussion to highlight the advantages of \system{} over state-of-the-art approaches. We then conduct a quantitative comparison to further verify the improvements of \system{} over directly-comparable prior work.

\textit{(1) Qualitative analysis.}
Prior VLA testing studies are divided into two categories: manually curated benchmarks, such as  LIBERO-Plus\cite{fei2025libero}, RobustVLA\cite{zhang2025robustvla}, Eva-VLA\cite{liu2025eva}, ViFailback~\cite{zeng2025diagnose}, and VLA-Arena~\cite{zhang2025vla}, and automated testing frameworks, such as VLATest~\cite{wang2025vlatest}. In contrast, \system{} alleviates the key limitations of both categories.

Compared with manually-curated benchmarks, which require substantial human effort and cover only limited configuration dimensions, \system{} automatically produces test inputs and thus avoids such labeling costs and dimensional restrictions, allowing it to discover new failure modes at scale. Since their evaluation is bound to a predefined input set, these benchmarks are not directly comparable with \system{} using metrics such as test coverage. A direct quantitative comparison is therefore not methodologically meaningful, so we discuss these benchmarks only qualitatively.

Compared with VLATest, whose random candidate generation ignores the selection of failure-revealing inputs and the reduction of inter-sample redundancy, \system{} employs a distance-aware selection strategy that prioritizes candidates far from the training distribution while pruning candidates that are mutually close in the latent space, yielding a more diverse and failure-revealing test set under the same budget. We next quantitatively verify these advantages against VLATest.

\textit{(2) Quantitative analysis.}
To compare the effectiveness of \system{} and VLATest in revealing failures, we conducted an experiment in which each tool generated 400 test cases for each VLA model, with 100 cases for each of \textit{Pick-up}, \textit{Move-to}, \textit{Put-on}, and \textit{Put-into}. Since VLATest supports only two of the three models, namely OpenVLA and RT-1, we run both \system{} and VLATest on these two models, resulting in 1,600 test cases in total, and compare their average test coverage and average failure detection rate. 

Table~\ref{tab:comparison} shows the comparison results between \system{} and VLATest. Under the same test budget, \system{} achieves higher failure detection and trajectory coverage rates than VLATest. For example, compared with VLATest, \system{} improves failure detection by 6.76\% and trajectory coverage by 18.93\%. These results indicate that \system{} outperforms the state-of-the-art tool in terms of both testing effectiveness and diversity.

\subsection{RQ2: Effectiveness in Repairing VLA Failures}
\label{sec:RQ2}

We evaluate the effectiveness of \system{} in leveraging detected failures to enhance VLA models. Moreover, we further analyze persistent failures that remain after repairing the VLA models.

Based on the set of failure trajectories obtained above, we enhance all VLA models by following the procedure described in Section~\ref{DataSynthesis}. In this manner, we obtain corresponding enhanced models for the three original VLA models, denoted as RT-1-X-\system{}, OpenVLA-\system{}, and $\pi_{0.5}$-\system{}.
We then generate 1200 newly-generated unseen test cases in the simulation environment to test these enhanced models, for evaluating their success rate improvement. Next, we calculate the total number of failures and the number of failures across different robotic manipulation tasks.
Subsequently, we compute the performance improvements of all three enhanced models.
We refer to test cases that still fail after model enhancement as non-repaired tests.
To mitigate the impact of non-determinism in the simulation environment, we repeat each non-repaired test five additional times.

\begin{table}[t]
\centering
\footnotesize
\caption{The test analysis of model enhanced by \system{}, for evaluating the effectiveness in repairing VLA's inefficiencies}
\label{tab:repar}
\setlength{\tabcolsep}{3pt}

\begin{tabular}{lcccccccc}
\toprule
\textbf{VLA Models} & \multicolumn{3}{c}{\textbf{All Tests}} & \multicolumn{4}{c}{\textbf{Failures in different tasks}} \\
\cmidrule(lr){2-4}\cmidrule(lr){5-8}
                   & Pass   & Fail   &Total    & Pick up   & Move to   & Put on  & Put in\\
\midrule
RT-1-X-\system{}      & 214    & 186    & 400     & 44        & 23        & 68      & 51      \\
OpenVLA-\system{}     & 233    & 167    & 400     & 41        & 19        & 69      & 38      \\
$\pi_{0.5}$-\system{} & 236    & 164    & 400     & 21        & 35        & 51      & 57      \\
 \hline
\textbf{Total}     & 683    & 517    & 1200    & 106       & 77        & 188     & 146       \\ 
\bottomrule
\end{tabular}


\end{table}

Table~\ref{tab:repar} shows the test results of the enhanced VLA models.
We compare these results with the performance of the original VLA models shown in Table~\ref{tab:overall}. As shown, \system{} achieves strong overall repair effectiveness, successfully enhance the success rate by 49.25\% ((683-92)/1200).

\para{Adaptation to Different Robotic Manipulation Tasks}
Overall, \system{} achieves substantial improvements across different robotic manipulation tasks, consistently improving the success rate by 37.00\% to 62.67\%. For example, on the \textit{Pick-up} and \textit{Move-to} tasks, \system{} enhances the success rate by 54.33\% ((194-31)/300) and 62.67\% ((223-35)/300), respectively. For the more complex \textit{Put-on} and \textit{Put-into} tasks, the performance enhancements are also pronounced, reaching 37.00\% ((112-1)/300) and 43.00\% ((154-25)/300), respectively. These results demonstrate that \system{} achieves strong failure-repair performance across robotic manipulation tasks with different difficulty levels.

\para{Adaptation to Different Models}
When applied to different VLA models, \system{} also maintains consistently-strong repair performance. Specifically, for monolithic VLA models such as RT-1 and OpenVLA, \system{} increases the success rate by 44.00\% ((214-38)/400) and 48.50\% ((233-39)/400), respectively. For the more complex hierarchical model $\pi_{0.5}$, \system{} achieves a higher success rate improvement of 55.25\% ((236-15)/400). These findings indicate that \system{} generalizes well to enhancing diverse VLA models, regardless of model architecture, reasoning capability, or model size.

\para{Analysis of Non-repaired Tests}
Furthermore, we manually inspected all non-repaired test cases. The inspection results indicate that most non-repaired cases are caused by anomalies introduced by the model architecture or training process. For example, a recurring issue is unexplained jitter during robotic manipulation, which shifts the target object's position and ultimately leads to task failure.
Another notable issue is erratic robot motion, which causes the camera to lose track of the target object and results in task failure.
In fact, such issues inherently require improvements to the model architecture and training process, and thus cannot be directly addressed by analysis tools such as ours.

\subsection{RQ3: Real-world Performance of Optimized VLA Models}
\label{sec:RQ3}

While \system{} has been shown to be effective in simulation, our primary goal in enhancing VLA models is to facilitate their transfer to real-world environments at low cost.

Given the high cost of real-world testing, we additionally run \system{} to generate 80 new unseen test inputs, corresponding to 20\% of the evaluation budget used in RQ1. We construct corresponding real-world test setups for these 80 new cases, with scene configurations (e.g., object categories, relative positions, task instructions, and scene layouts) consistent with those in the simulation environment. To reduce experimental overhead, we mainly conduct real-world evaluations on the $\pi_{0.5}$ model, as RQ1 and RQ2 indicate that \system{} shows strong generalizability and effectiveness across model types.

\begin{table}[t]
\centering
\footnotesize
\caption{Performance of original model and model optimized with in-simulation synthetic data on the real world data}
\label{tab:sim}
\setlength{\tabcolsep}{3.5pt}

\begin{tabular}{lcccccccc}
\toprule
\textbf{VLA Models} & \multicolumn{3}{c}{\textbf{All Tests}} & \multicolumn{4}{c}{\textbf{Failures in different tasks}} \\
\cmidrule(lr){2-4}\cmidrule(lr){5-8}
                   & Pass   & Fail   &Total    & Pick up   & Move to   & Put on  & Put in\\
\midrule
$\pi_{0.5}$        & 0    & 80    & 80     & 20        & 20        & 20      & 20      \\
$\pi_{0.5}$-\system{} & 46    & 34    & 80     & 5        & 7        & 10      & 12      \\
\bottomrule
\end{tabular}
\end{table}

Our hypothesis is that when a model is enhanced using a test-generated dataset collected in simulation, the enhanced model remains effective in real-world scenarios. To verify this hypothesis, we evaluate two models on real-world tasks: (1) the original $\pi_{0.5}$ model and (2) $\pi_{0.5}$-\system{}, which is obtained by model enhancement using the test-generated dataset from simulation.
All evaluations are conducted in real-world scenes.

Table~\ref{tab:sim} reports the results of $\pi_{0.5}$ and $\pi_{0.5}$-\system{} on the real-world tasks.
As shown, the test tasks generated by \system{} are also effective in real-world settings, with 100\% (80/80) of them resulting in VLA failures. Moreover, \system{} remains effective across manipulation tasks with different difficulty levels, achieving a 100\% failure rate.
Regarding model repair, the enhanced model $\pi_{0.5}$-\system{} remains effective on real-world tasks and significantly outperforms the original $\pi_{0.5}$ model. Compared with $\pi_{0.5}$, the enhanced model $\pi_{0.5}$-\system{} achieves 46 successful cases, corresponding to an improving success rate of 57.50\% ((46-0)/80) in real-world scenes.
These findings indicate that \system{} maintains strong performance on new real-world tasks and identifies and repairs failures in a general and effective manner. In particular, simulation-repaired VLA models remain effective in real-world settings, highlighting a promising low-cost direction for improving VLA models.

\subsection{Discussion and Limitation}
\label{sec:discuss}

\system{} has the following advantages. (1) As evidenced by the evaluation results, \system{} effectively reveals VLA failures and leverages the identified failures to enhance the model. (2) \system{} integrates our proposed distance-aware test selection strategy, which improves both the effectiveness and diversity of the generated test cases. This strategy mitigates the limitations of prior work. 
(3) For failure trajectories, \system{} uses a VLM-based agent to generate successfully-executed trajectories, which are automatically annotated as supervisory samples to fine-tune the original VLA model for model enhancement.
These designs provide a closed loop from failure discovery to model repair, which remain insufficiently supported in prior work.

\para{Threats to Validity} One external threat to validity is the reliance on a predefined repair-skill library.
However, this reliance does not limit the applicability of \system{} for the following reasons. (1) The current skill library is designed around the four targeted manipulation tasks and covers key repair primitives, including object relocation, grasp correction, pose adjustment, placement correction, and success verification. These skills collectively suffice to recover from the failures observed in our evaluation. (2) Rather than hard-coding an end-to-end procedure for each task, \system{} decomposes the repair process into reusable primitive-level skills. Different tasks are handled by recomposing the same primitives with task-specific parameters, which gives the design strong generality across manipulation scenarios. (3) The skill library is organized in a backward-compatible manner, so adding a new skill requires only lightweight effort.

One internal threat to validity is the manual analysis involved in the evaluation, which may introduce subjective bias.
To mitigate this threat, we standardize the manual analysis procedure into the following steps.
First, we invite six researchers with at least two years of domain research experience and divide them into three groups. Two groups serve as annotators, while the remaining group serves as referee experts with at least four years of domain research experience.
Second, each annotator group conducts manual analysis independently.
Third, within each annotator group, researchers cross-validate their annotations and reach consensus on the classification outcomes through discussion. We also compute Cohen's kappa coefficient to quantify inter-annotator agreement. If Cohen's kappa falls below a predefined threshold, the referee experts are invited to intervene and make the final decision.
In addition, we compute the mean Cohen's kappa for the overall manual analysis process and obtain a value of 0.764, indicating substantial agreement among the researchers.

\section{Related work}
\label{sec:relatedwork}

\para{Benchmarking and Evaluating VLA Models}
A growing number of studies have focused on evaluating VLA models across various properties.
Prior studies, including Eva-VLA~\cite{liu2025eva}, LIBERO-Plus~\cite{fei2025libero}, RobustVLA~\cite{zhang2025robustvla}, and VLATest~\cite{wang2025vlatest}, analyze the robustness of VLA models by introducing perturbations or employing randomized testing. Zhang et al. construct VLA-Arena, a benchmark dataset for evaluating VLA models along four key dimensions, including safety, distractors, extrapolation, and long-horizon tasks~\cite{zhang2025vla}. In addition, Valle et al. conduct a quantitative analysis of the uncertainty and quality of VLA models~\cite{valle2025evaluating}.
ViFailback~\cite{zeng2025diagnose} is designed to diagnose robotic manipulation failures and provide both textual and visual correction guidance.
Compared with these methods, we develop a more general VLA testing framework that covers representative VLA failure types and generates diverse test cases to more thoroughly explore the capability boundaries of VLA models.

\para{Deep Neural Network Repair}
Prior studies have investigated repairing and improving deep neural networks (DNNs) through data augmentation from different perspectives~\cite{zohdinasab2023deepatash,wang2021robot,dong2019there,harel2020neuron,attaoui2025search,humeniuk2024simulation,du2019deepstellar,li2021testing,yu2021deeprepair,gao2020fuzz}. Specifically, DESIGNATE~\cite{attaoui2025search} uses generative adversarial networks (GANs) to map simulation-generated data into more realistic images and then retrains the DNN accordingly. TACTIC~\cite{li2021testing} performs image-to-image translation under diverse environmental conditions to synthesize additional training data. DeepRepair~\cite{yu2021deeprepair} adopts a style-transfer-based augmentation mechanism, while SENSEI~\cite{gao2020fuzz} further proposes a search-based augmentation strategy.
However, these approaches typically achieve limited performance improvements because they ignore the critical semantics inherent in robotic manipulation. In contrast, \system{} captures both planning and action semantics and repairs VLA models by correcting semantic errors in failure trajectories.

\section{Conclusion}
\label{sec:conclusion}

This paper proposes \system{}, a novel framework for revealing VLA failures and enhancing VLA models through failure trajectories. We conduct a comprehensive evaluation of \system{} across three experiments, including failure detection, model repair, and real-world performance.
The evaluation results show that \system{} significantly outperforms existing tools in revealing VLA failures, with 92.33\% of test inputs resulting in VLA failures and test coverage improved by 18.93\%.
By performing agent-based repair of VLA models through failure trajectories, \system{} enhances the overall success rate by 49.25\%. 
Finally, sim-to-real transfer experiments indicate that when a model is enhanced by using a test-generated dataset collected in simulation, the enhanced model remains effective in real-world scenarios. Overall, these results pinpoint a promising and low-cost direction for VLA model enhancement.

\section{Data availability}

Currently, the artifact of \system{} and other online materials are available in this anonymous repository (\url{https://github.com/scuama/VLAMotor}).


\normalem

{
    \bibliographystyle{ieeetr}
    \bibliography{ref}

@article{shao2025large,
  title={Large vlm-based vision-language-action models for robotic manipulation: A survey},
  author={Shao, Rui and Li, Wei and Zhang, Lingsen and Zhang, Renshan and Liu, Zhiyang and Chen, Ran and Nie, Liqiang},
  journal={arXiv preprint arXiv:2508.13073},
  year={2025}
}

@inproceedings{cohen2020detecting,
  title={Detecting adversarial samples using influence functions and nearest neighbors},
  author={Cohen, Gilad and Sapiro, Guillermo and Giryes, Raja},
  booktitle={Proceedings of the IEEE/CVF conference on computer vision and pattern recognition},
  pages={14453--14462},
  year={2020}
}

@article{karpusha2020calibrated,
  title={Calibrated neighborhood aware confidence measure for deep metric learning},
  author={Karpusha, Maryna and Yun, Sunghee and Fehervari, Istvan},
  journal={arXiv preprint arXiv:2006.04935},
  year={2020}
}

@article{papernot2018deep,
  title={Deep k-nearest neighbors: Towards confident, interpretable and robust deep learning},
  author={Papernot, Nicolas and McDaniel, Patrick},
  journal={arXiv preprint arXiv:1803.04765},
  year={2018}
}

@inproceedings{romano2006exploring,
  title={Exploring methods for evaluating group differences on the NSSE and other surveys: Are the t-test and Cohen’sd indices the most appropriate choices},
  author={Romano, Jeanine and Kromrey, Jeffrey D and Coraggio, Jesse and Skowronek, Jeff and Devine, Linda},
  booktitle={annual meeting of the Southern Association for Institutional Research},
  volume={14},
  year={2006},
  organization={Citeseer}
}

@inproceedings{frosst2019analyzing,
  title={Analyzing and improving representations with the soft nearest neighbor loss},
  author={Frosst, Nicholas and Papernot, Nicolas and Hinton, Geoffrey},
  booktitle={International conference on machine learning},
  pages={2012--2020},
  year={2019},
  organization={PMLR}
}

@article{jiang2018trust,
  title={To trust or not to trust a classifier},
  author={Jiang, Heinrich and Kim, Been and Guan, Melody and Gupta, Maya},
  journal={Advances in neural information processing systems},
  volume={31},
  year={2018}
}

@article{wang2025element,
  title={Element-Based Automated DNN Repair with Fine-Tuned Masked Language Model},
  author={Wang, Xu and Zhang, Mingming and Meng, Xiangxin and Zhang, Jian and Liu, Yang and Hu, Chunming},
  journal={Proceedings of the ACM on Software Engineering},
  volume={2},
  number={FSE},
  pages={106--129},
  year={2025},
  publisher={ACM New York, NY, USA}
}

@inproceedings{li2022hybridrepair,
  title={HybridRepair: towards annotation-efficient repair for deep learning models},
  author={Li, Yu and Chen, Muxi and Xu, Qiang},
  booktitle={Proceedings of the 31st ACM SIGSOFT International Symposium on Software Testing and Analysis},
  pages={227--238},
  year={2022}
}

@article{attaoui2025search,
  title={Search-based dnn testing and retraining with gan-enhanced simulations},
  author={Attaoui, Mohammed Oualid and Pastore, Fabrizio and Briand, Lionel C},
  journal={IEEE Transactions on Software Engineering},
  year={2025},
  publisher={IEEE}
}

@article{attaoui2023black,
  title={Black-box safety analysis and retraining of dnns based on feature extraction and clustering},
  author={Attaoui, Mohammed and Fahmy, Hazem and Pastore, Fabrizio and Briand, Lionel},
  journal={ACM Transactions on Software Engineering and Methodology},
  volume={32},
  number={3},
  pages={1--40},
  year={2023},
  publisher={ACM New York, NY}
}

@inproceedings{li2024distance,
  title={Distance-aware test input selection for deep neural networks},
  author={Li, Zhong and Xu, Zhengfeng and Ji, Ruihua and Pan, Minxue and Zhang, Tian and Wang, Linzhang and Li, Xuandong},
  booktitle={Proceedings of the 33rd ACM SIGSOFT International Symposium on Software Testing and Analysis},
  pages={248--260},
  year={2024}
}

@article{guan2025efficient,
  title={Efficient vision-language-action models for embodied manipulation: A systematic survey},
  author={Guan, Weifan and Hu, Qinghao and Li, Aosheng and Cheng, Jian},
  journal={arXiv preprint arXiv:2510.17111},
  year={2025}
}

@article{dzedzickis2021advanced,
  title={Advanced applications of industrial robotics: New trends and possibilities},
  author={Dzedzickis, Andrius and Suba{\v{c}}i{\=u}t{\.e}-{\v{Z}}emaitien{\.e}, Jurga and {\v{S}}utinys, Ernestas and Samukait{\.e}-Bubnien{\.e}, Urt{\.e} and Bu{\v{c}}inskas, Vytautas},
  journal={Applied Sciences},
  volume={12},
  number={1},
  pages={135},
  year={2021},
  publisher={MDPI}
}

@article{holland2021service,
  title={Service robots in the healthcare sector},
  author={Holland, Jane and Kingston, Liz and McCarthy, Conor and Armstrong, Eddie and O’Dwyer, Peter and Merz, Fionn and McConnell, Mark},
  journal={Robotics},
  volume={10},
  number={1},
  pages={47},
  year={2021},
  publisher={MDPI}
}

@incollection{dhaliwal2020rise,
  title={The rise of automation and robotics in warehouse management},
  author={Dhaliwal, Amandeep},
  booktitle={Transforming management using artificial intelligence techniques},
  pages={63--72},
  year={2020},
  publisher={CRC Press}
}

@inproceedings{humeniuk2024simulation,
  title={In-Simulation Testing of Deep Learning Vision Models in Autonomous Robotic Manipulators},
  author={Humeniuk, Dmytro and Ben Braiek, Houssem and Reid, Thomas and Khomh, Foutse},
  booktitle={Proceedings of the 39th IEEE/ACM International Conference on Automated Software Engineering},
  pages={2187--2198},
  year={2024}
}

@article{wang2025vlatest,
  title={VLATest: Testing and Evaluating Vision-Language-Action Models for Robotic Manipulation},
  author={Wang, Zhijie and Zhou, Zhehua and Song, Jiayang and Huang, Yuheng and Shu, Zhan and Ma, Lei},
  journal={Proceedings of the ACM on Software Engineering},
  volume={2},
  number={FSE},
  pages={1615--1638},
  year={2025},
  publisher={ACM New York, NY, USA}
}

@article{fei2025libero,
  title={Libero-plus: In-depth robustness analysis of vision-language-action models},
  author={Fei, Senyu and Wang, Siyin and Shi, Junhao and Dai, Zihao and Cai, Jikun and Qian, Pengfang and Ji, Li and He, Xinzhe and Zhang, Shiduo and Fei, Zhaoye and others},
  journal={arXiv preprint arXiv:2510.13626},
  year={2025}
}

@article{zhang2025robustvla,
  title={RobustVLA: Robustness-Aware Reinforcement Post-Training for Vision-Language-Action Models},
  author={Zhang, Hongyin and Zhang, Shuo and Jin, Junxi and Zeng, Qixin and Li, Runze and Wang, Donglin},
  journal={arXiv preprint arXiv:2511.01331},
  year={2025}
}

@article{liu2025eva,
  title={Eva-VLA: Evaluating Vision-Language-Action Models' Robustness Under Real-World Physical Variations},
  author={Liu, Hanqing and Long, Jiahuan and Wu, Junqi and Hou, Jiacheng and Tang, Huili and Jiang, Tingsong and Zhou, Weien and Yao, Wen},
  journal={arXiv preprint arXiv:2509.18953},
  year={2025}
}

@article{xu2024survey,
  title={A survey on robotics with foundation models: toward embodied ai},
  author={Xu, Zhiyuan and Wu, Kun and Wen, Junjie and Li, Jinming and Liu, Ning and Che, Zhengping and Tang, Jian},
  journal={arXiv preprint arXiv:2402.02385},
  year={2024}
}

@article{ma2024survey,
  title={A survey on vision-language-action models for embodied ai},
  author={Ma, Yueen and Song, Zixing and Zhuang, Yuzheng and Hao, Jianye and King, Irwin},
  journal={arXiv preprint arXiv:2405.14093},
  year={2024}
}

@article{zhang2025generative,
  title={Generative artificial intelligence in robotic manipulation: A survey},
  author={Zhang, Kun and Yun, Peng and Cen, Jun and Cai, Junhao and Zhu, Didi and Yuan, Hangjie and Zhao, Chao and Feng, Tao and Wang, Michael Yu and Chen, Qifeng and others},
  journal={arXiv preprint arXiv:2503.03464},
  year={2025}
}

@article{liu2025aligning,
  title={Aligning cyber space with physical world: A comprehensive survey on embodied ai},
  author={Liu, Yang and Chen, Weixing and Bai, Yongjie and Liang, Xiaodan and Li, Guanbin and Gao, Wen and Lin, Liang},
  journal={IEEE/ASME Transactions on Mechatronics},
  year={2025},
  publisher={IEEE}
}

@incollection{goel2019robotics,
  title={Robotics and industry 4.0},
  author={Goel, Ruchi and Gupta, Pooja},
  booktitle={A roadmap to industry 4.0: Smart production, sharp business and sustainable development},
  pages={157--169},
  year={2019},
  publisher={Springer}
}

@inproceedings{o2024open,
  title={Open x-embodiment: Robotic learning datasets and rt-x models: Open x-embodiment collaboration 0},
  author={O’Neill, Abby and Rehman, Abdul and Maddukuri, Abhiram and Gupta, Abhishek and Padalkar, Abhishek and Lee, Abraham and Pooley, Acorn and Gupta, Agrim and Mandlekar, Ajay and Jain, Ajinkya and others},
  booktitle={2024 IEEE International Conference on Robotics and Automation (ICRA)},
  pages={6892--6903},
  year={2024},
  organization={IEEE}
}

@article{kim2024openvla,
  title={Openvla: An open-source vision-language-action model},
  author={Kim, Moo Jin and Pertsch, Karl and Karamcheti, Siddharth and Xiao, Ted and Balakrishna, Ashwin and Nair, Suraj and Rafailov, Rafael and Foster, Ethan and Lam, Grace and Sanketi, Pannag and others},
  journal={arXiv preprint arXiv:2406.09246},
  year={2024}
}

@article{intelligence2025pi_,
  title={ A Vision-Language-Action Model with Open-World Generalization},
  author={Intelligence, Physical and Black, Kevin and Brown, Noah and Darpinian, James and Dhabalia, Karan and Driess, Danny and Esmail, Adnan and Equi, Michael and Finn, Chelsea and Fusai, Niccolo and others},
  journal={arXiv preprint arXiv:2504.16054},
  year={2025}
}

@inproceedings{gr00tn1_2025,
  archivePrefix = {arxiv},
  eprint     = {2503.14734},
  title      = {{GR00T} {N1}: An Open Foundation Model for Generalist Humanoid Robots},
  author     = {NVIDIA and Johan Bjorck and Fernando Castañeda, Nikita Cherniadev and Xingye Da and Runyu Ding and Linxi "Jim" Fan and Yu Fang and Dieter Fox and Fengyuan Hu and Spencer Huang and Joel Jang and Zhenyu Jiang and Jan Kautz and Kaushil Kundalia and Lawrence Lao and Zhiqi Li and Zongyu Lin and Kevin Lin and Guilin Liu and Edith Llontop and Loic Magne and Ajay Mandlekar and Avnish Narayan and Soroush Nasiriany and Scott Reed and You Liang Tan and Guanzhi Wang and Zu Wang and Jing Wang and Qi Wang and Jiannan Xiang and Yuqi Xie and Yinzhen Xu and Zhenjia Xu and Seonghyeon Ye and Zhiding Yu and Ao Zhang and Hao Zhang and Yizhou Zhao and Ruijie Zheng and Yuke Zhu},
  month      = {March},
  year       = {2025},
  booktitle  = {ArXiv Preprint},
}

@inproceedings{du2019deepstellar,
  title={Deepstellar: Model-based quantitative analysis of stateful deep learning systems},
  author={Du, Xiaoning and Xie, Xiaofei and Li, Yi and Ma, Lei and Liu, Yang and Zhao, Jianjun},
  booktitle={Proceedings of the 2019 27th ACM joint meeting on European software engineering conference and symposium on the foundations of software engineering},
  pages={477--487},
  year={2019}
}

@article{wu2025human,
  title={From human memory to ai memory: A survey on memory mechanisms in the era of llms},
  author={Wu, Yaxiong and Liang, Sheng and Zhang, Chen and Wang, Yichao and Zhang, Yongyue and Guo, Huifeng and Tang, Ruiming and Liu, Yong},
  journal={arXiv preprint arXiv:2504.15965},
  year={2025}
}

@article{sarch2024vlm,
  title={Vlm agents generate their own memories: Distilling experience into embodied programs of thought},
  author={Sarch, Gabriel and Jang, Lawrence and Tarr, Michael and Cohen, William W and Marino, Kenneth and Fragkiadaki, Katerina},
  journal={Advances in Neural Information Processing Systems},
  volume={37},
  pages={75942--75985},
  year={2024}
}

@article{zhang2025vla,
  title={VLA-Arena: An Open-Source Framework for Benchmarking Vision-Language-Action Models},
  author={Zhang, Borong and Li, Jiahao and Shen, Jiachen and Cai, Yishuai and Zhang, Yuhao and Chen, Yuanpei and Dai, Juntao and Ji, Jiaming and Yang, Yaodong},
  journal={arXiv preprint arXiv:2512.22539},
  year={2025}
}

@article{valle2025evaluating,
  title={Evaluating uncertainty and quality of visual language action-enabled robots},
  author={Valle, Pablo and Lu, Chengjie and Ali, Shaukat and Arrieta, Aitor},
  journal={arXiv preprint arXiv:2507.17049},
  year={2025}
}

@article{zeng2025diagnose,
  title={Diagnose, Correct, and Learn from Manipulation Failures via Visual Symbols},
  author={Zeng, Xianchao and Zhou, Xinyu and Li, Youcheng and Shi, Jiayou and Li, Tianle and Chen, Liangming and Ren, Lei and Li, Yong-Lu},
  journal={arXiv preprint arXiv:2512.02787},
  year={2025}
}

@article{yu2021deeprepair,
  title={Deeprepair: Style-guided repairing for deep neural networks in the real-world operational environment},
  author={Yu, Bing and Qi, Hua and Guo, Qing and Juefei-Xu, Felix and Xie, Xiaofei and Ma, Lei and Zhao, Jianjun},
  journal={IEEE Transactions on Reliability},
  volume={71},
  number={4},
  pages={1401--1416},
  year={2021},
  publisher={IEEE}
}

@inproceedings{gao2020fuzz,
  title={Fuzz testing based data augmentation to improve robustness of deep neural networks},
  author={Gao, Xiang and Saha, Ripon K and Prasad, Mukul R and Roychoudhury, Abhik},
  booktitle={Proceedings of the acm/ieee 42nd international conference on software engineering},
  pages={1147--1158},
  year={2020}
}

@inproceedings{li2021testing,
  title={Testing dnn-based autonomous driving systems under critical environmental conditions},
  author={Li, Zhong and Pan, Minxue and Zhang, Tian and Li, Xuandong},
  booktitle={International conference on machine learning},
  pages={6471--6482},
  year={2021},
  organization={PMLR}
}

@inproceedings{zohdinasab2023deepatash,
  title={Deepatash: Focused test generation for deep learning systems},
  author={Zohdinasab, Tahereh and Riccio, Vincenzo and Tonella, Paolo},
  booktitle={Proceedings of the 32nd ACM SIGSOFT international symposium on software testing and analysis},
  pages={954--966},
  year={2023}
}

@article{dong2019there,
  title={There is limited correlation between coverage and robustness for deep neural networks},
  author={Dong, Yizhen and Zhang, Peixin and Wang, Jingyi and Liu, Shuang and Sun, Jun and Hao, Jianye and Wang, Xinyu and Wang, Li and Dong, Jin Song and Ting, Dai},
  journal={arXiv preprint arXiv:1911.05904},
  year={2019}
}

@inproceedings{harel2020neuron,
  title={Is neuron coverage a meaningful measure for testing deep neural networks?},
  author={Harel-Canada, Fabrice and Wang, Lingxiao and Gulzar, Muhammad Ali and Gu, Quanquan and Kim, Miryung},
  booktitle={Proceedings of the 28th ACM Joint Meeting on European Software Engineering Conference and Symposium on the Foundations of Software Engineering},
  pages={851--862},
  year={2020}
}

@inproceedings{wang2021robot,
  title={Robot: Robustness-oriented testing for deep learning systems},
  author={Wang, Jingyi and Chen, Jialuo and Sun, Youcheng and Ma, Xingjun and Wang, Dongxia and Sun, Jun and Cheng, Peng},
  booktitle={2021 IEEE/ACM 43rd International Conference on Software Engineering (ICSE)},
  pages={300--311},
  year={2021},
  organization={IEEE}
}
}




\end{document}